\documentclass{article}
\usepackage{ijcai22}

\usepackage{times}
\usepackage{soul}
\usepackage{url}
\usepackage[hidelinks]{hyperref}
\usepackage[utf8]{inputenc}
\usepackage[small]{caption}
\usepackage{graphicx}
\usepackage{amsmath}
\usepackage{amsfonts}
\usepackage{amsthm}
\usepackage{booktabs}
\usepackage{algorithm}
\usepackage{algorithmic}
\usepackage{multirow}
\usepackage{xcolor}
\usepackage{subfigure}
\title{Leveraging Intrinsic Gradient Information for Further Training of Differentiable Machine Learning Models}

\author{
Chris McDonagh$^1$\and
Xi Chen$^1$\footnote{Corresponding author.}
\affiliations
$^1$Department of Computer Science, University of Bath, Bath, BA2 7PB, UK\\
\emails
\{cm2042, xc841\}@bath.ac.uk
}

\begin{document}

\maketitle

\begin{abstract}
Designing models that produce accurate predictions is the fundamental objective of machine learning (ML). This work presents methods demonstrating that when the derivatives of target variables (outputs) with respect to inputs can be extracted from processes of interest, e.g., neural networks (NN) based surrogate models, they can be leveraged to further improve the accuracy of differentiable ML models. This paper generalises the idea and provides practical methodologies that can be used to leverage gradient information (GI) across a variety of applications including: (1) Improving the performance of generative adversarial networks (GANs); (2) efficiently tuning NN model complexity; (3) regularising linear regressions. Numerical results show that GI can effective enhance ML models with existing datasets, demonstrating its value for a variety of applications.
\end{abstract}

\section{Introduction}

Designing models that can accurately make predictions is the fundamental objective of machine learning. The idea of using gradient information (GI) to further improve ML model fit in the context of computer simulator based systems was firstly explored in~\cite{Brehmer5242}, in which the authors investigated latent variable estimation problems for computationally expensive computer simulators and demonstrated how the derivative of joint probability from stochastic simulators can be used to improve the accuracy of NN conditional density estimators. GI can be used in conjunction with a classification method referred to as the Likelihood Ratio Trick (LRT) (see ~\cite{tran2017hierarchical}, ~\cite{gutmann2018likelihood} and ~\cite{Brehmer5242} for more details) to improve the estimation of unknown parameter probability densities using NN models. 





In this paper, we further generalise the usage of GI to a broader range of systems/processes where the derivatives of outputs with respects to inputs can be explicitly computed. We aim to highlight to practitioners that more information to train models may be available by way of GI. In light of GI, we develop novel and practical methodologies for the following three types of applications: 
\begin{itemize}
    \item We present a novel GAN architecture that utilises GI to further improve the performance of generative NN models
	\item We introduce a novel metric that can be utilised in a hyper-parameter optimisation pipeline that provides an indicator of an upper bound to NN model complexity
	\item We propose an alternative regularisation method for linear regression problems (using ridge regression as an example) that outperforms conventional regularisation over varying training sample sizes by utilising GI
\end{itemize}

In the rest of this paper, Section \ref{grads_NN_training} formulates the GI idea under a supervised learning setting. The proposed GI assisted methodologies are presented between Section 3 to 5, and followed by a conclusion in Section \ref{Con}. 


\section{Further training of supervised learning models with GI}
\label{grads_NN_training}

Consider a real-world process $P$ that takes inputs $X$ to produce outputs $Z$:
\begin{equation}
\mathbf{Z} = P(\mathbf{X}).
\end{equation}
Supervised learning aims to design and train a model to emulate this process $\hat{P}$:
\begin{equation}
	\mathbf{\hat{z}} = \hat{P}_{\varphi}(\mathbf{x}).
\end{equation}
This requires collecting data on the independent variable $\mathbf{z}\subseteq \mathbf{Z}$ and dependent variables $\mathbf{x}\subseteq \mathbf{X}$ for which data exist. The parameters of the selected model can then be trained using training data with $N$ samples $\{\mathbf{x}_i, \mathbf{z}_i\}_{i=1}^N$.


\begin{figure*}[t]
	\centering
	\hbox{\hspace{-7.5em}\includegraphics[width=22cm]{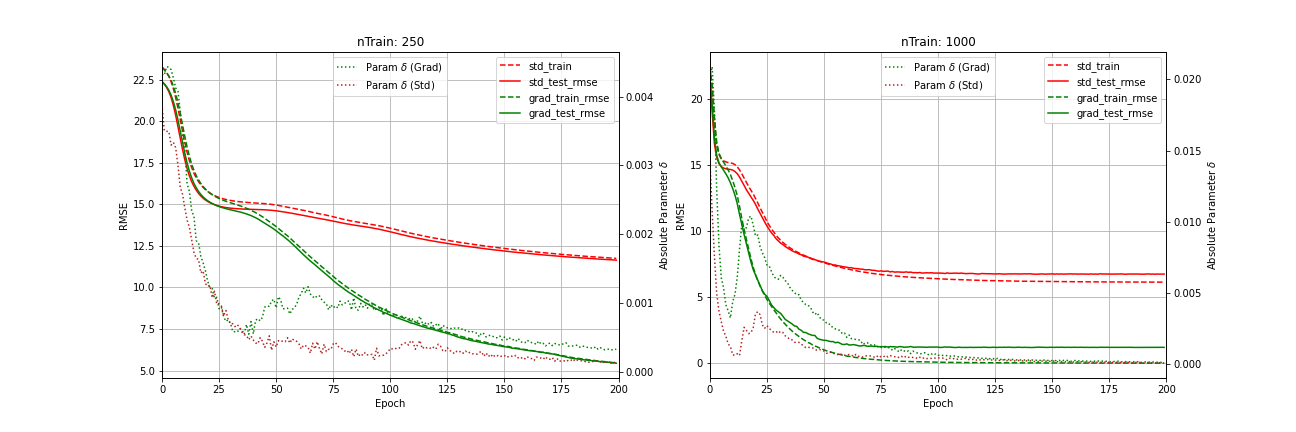}}
	\caption{Train and test root mean-squared error (RMSE) of neural networks trained with and without gradients. Left figure uses smaller training sample size of 250, right uses larger sample of 1000. Model accuracy is shown on left axes and absolute parameter delta is shown on right axes. Results are averaged over 5 experimental runs.} 
	\label{tp_epochs}
\end{figure*}

\subsection{Why is GI valuable?}
\label{grad_val}
In supervised learning, if the derivatives of process outputs with respect to their inputs $\delta \mathbf{z_i}/\delta \mathbf{x_i}$ can be extracted, it will automatically expand the training dataset:
\begin{align*}
    \{\mathbf{x}_i, \mathbf{z}_i\}_{i=1}^N \rightarrow \{\mathbf{x}_i, \mathbf{z}_i, \frac{\delta \mathbf{z_i}}{\delta \mathbf{x_i}}\}_{i=1}^N
\end{align*}
These derivatives of process outputs are referred to throughout this paper as 'Gradient Information' (GI).

Consider standard back-propagation training of a feed-forward NN model: 
\begin{equation}
    \label{NN_update}
	\theta_i^+ = \theta_i - \eta \frac{\delta E}{\delta \theta_i},
\end{equation}
where $\theta_i$ is the $i$th parameter in network of given complexity and $\eta$ represents a learning rate. $E$ represents some error function which, using the training data $\{\mathbf{x}_i, \mathbf{z}_i\}_{i=1}^N$, defines how 'good' a model is. This standard feed-forward NN is differentiable with respect to its inputs and it learns to emulate $P$ via these parameter updates, which directly target the loss function $E$. Practitioners can therefore impose that not only should a network's outputs emulate target variables, but also that the derivative of network outputs with respect to inputs $\delta \mathbf{\hat{z}}/\delta \mathbf{x}$ should match the derivative of true outputs with respect to inputs $\delta \mathbf{z_i}/\delta \mathbf{x_i}$.

To implement this, an additional error function, termed as $E_{grad}$, is required to evaluate the accuracy between the NN derivatives and the true process derivatives. The parameter update mechanism therefore becomes:
\begin{equation}
    \label{NN_update_w_grad}
	\theta_i^+ = \theta_i - \alpha \eta \frac{\delta E}{\delta \theta_i} - (1-\alpha) \eta \frac{\delta E_{grad}}{\delta \theta_i},
\end{equation}
where $\alpha \in [0,1]$ is a weight coefficient. This can be critical when the magnitude of the target loss function $E$ and gradient loss function $E_{grad}$ differ materially. 

When trained with GI, Eq~\ref{NN_update_w_grad} shows that the absolute magnitude of each parameter update must be larger than when GI is not included\footnote{Note that a parameter update can be smaller when GI is included as the \textit{error that results from gradient fitting} can effectively offset the \textit{error that results from target fitting}, however the absolute error must be larger, despite their respective signs causing offsetting.}. Under the assumption that on average, a parameter update is beneficial for a model, which implies that model training generally is sensible (e.g. over-fitting is avoided), then the relative increase in the magnitude of parameter updates that result from the GI represents its marginal value. 

\subsection{An illustrative feed-forward NN example}


To illustrate this, consider the following toy function:
\begin{equation}
    \label{toy_func}
	\mathbf{Z} = 2\mathbf{X}^2 + \epsilon
\end{equation}
where $\epsilon \sim N(0, 1)$. To emulate this function, we design a feed-forward NN of sufficient complexity to avoid underfitting. $N$ samples of synthetic data, including GI, are then generated from the toy function and used to train the NN. Figure \ref{tp_epochs} illustrates this process, showing how the average update to NN parameters is larger each epoch when the network is trained with gradients. This in turn helps the gradient trained network outperform its conventionally trained counterpart. 

A key feature of the value of GI is that it reduces as the training sample size $N$ increases. This is shown in Figure \ref{tp_epochs}. With a relatively small training sample size of 250, the persistence in the relatively larger parameter updates per epoch for the gradient trained network drives out performance. In contrast, with the relatively larger training sample size of 1000, the delta of parameter changes per epoch converges sooner, due to the loss defined in equation \ref{NN_update_w_grad} reducing as a result of increased training sample size causing improved gradient fitting. 


\section{Gradient trained GAN}
\label{gan_grads}

Extending the insight of GI, we develop a novel gradient trained GAN architecture to further improve the performance of all GAN variants. GANs~\cite{goodfellow2014generative} are deep neural net architectures that are typically composed of a pair of competing NNs: a generator net $G$ and a discriminator net $D$. The models are trained by alternately optimising two objective functions so that the $G$ network learns to produce samples resembling real images, and the $D$ network learns to better discriminate between real and fake data.

\begin{figure}[!ht]
\centering
\includegraphics[width=5cm]{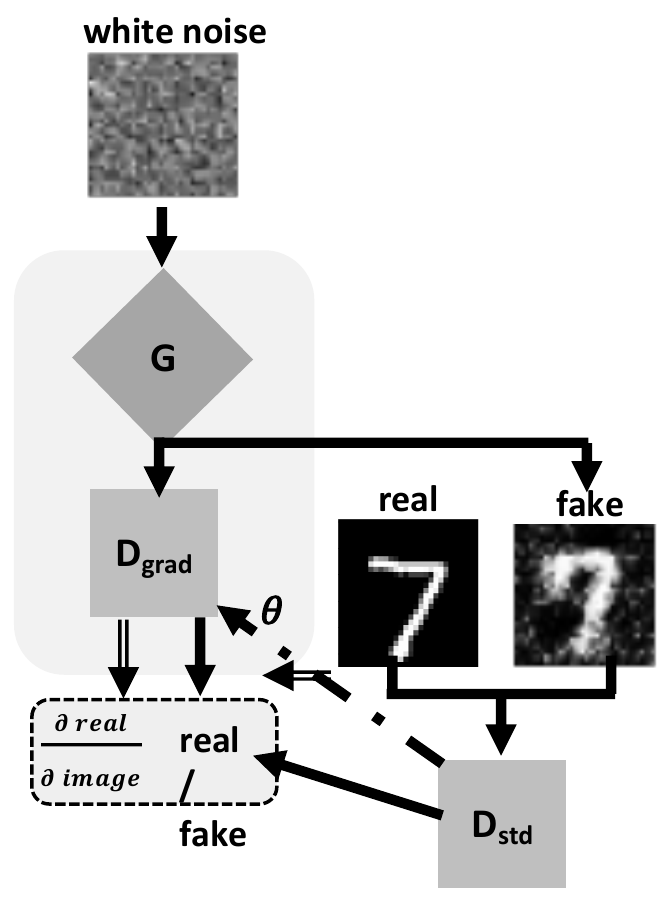}
\caption{Architecture of gradient trained GAN.}
\label{gan_grad_arch}
\end{figure}

Training GANs with gradients requires modifications to the conventional GAN architecture and training procedure, as shown in Figure \ref{gan_grad_arch}. First, two discriminators are used (rather than one) to facilitate the mechanics of gradient training. One discriminator $D_{std}$ is conventional in that it aims to classify images as real or fake without using GI. In each iteration, however, $D_{std}$ passes its updated parameters to a gradient discriminator $D_{grad}$. $D_{grad}$ then trains the $G$ network to produce images that it not only classifies as real (i.e., output from $D_{std}$), but also give 'real' derivatives of outputs with respect to inputs. The full pseudo-code is described in Algorithm \ref{alg:gan_grad}.

The loss of gradient trained GAN can be expressed by the following minmax objective function, where $G$ network tries to minimise it while the $D$ networks tries to maximise it:
\begin{align}
    &\underset{G}{\min} \underset{D}{\max} L(D, G) = \mathbb{E}_{Z}[\log D_{std/grad}(Z)] + \\ \nonumber
    &\mathbb{E}_{\epsilon}[\log (1 - D_{std/grad}(G(\epsilon)))] + \\ \nonumber
    &\gamma [D_{grad}(G(\epsilon)) - D_{grad}(Z)]^2, 
\end{align}
where $L(\cdot)$ represents the overall loss, and $\gamma$ is a pre-defined weight coefficient for gradient component. Due to weight passing between $D_{std}$ and $D_{grad}$ illustrated in Figure \ref{gan_grad_arch} and described in Algorithm \ref{alg:gan_grad}.6, they are effectively equivalent in the binary cross-entropy portion of the loss function. It is the final, mean-squared error term that leverages the gradient output of $D_{grad}$, expanding the loss function and regularising the updates to $G$.

\begin{figure}[!ht]
\includegraphics[width=8cm]{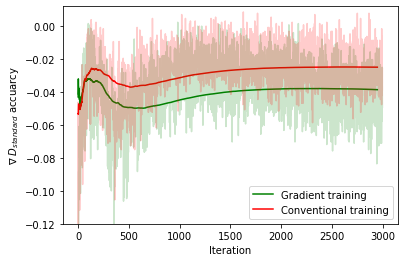}
\caption{The delta in $D_{std}$ accuracy, computed each iteration, for generators trained with and without GI. The negative delta for both series indicates generators produce a higher proportion of images that fool the discriminator after each training iteration. The rolling average of both series are plotted with solid line. Results use average of 10 experimental trials.}
\label{gan_eval}
\end{figure}

To demonstrate the value of GI for training GANs, standard GAN code is adapted to leverage GI for generating hand written digit images from the MNIST data set~\cite{deng2012mnist}. To evaluate the relative performance of a gradient trained GAN compared to a conventionally trained equivalent, two metrics are presented. First, the \textit{change in the accuracy} of $D_{std}$ is computed. Each iteration, the $G$ network is trained to fool an improved discriminator. Therefore by comparing the change in the accuracy of $D_{std}$ that results before and after training $G$, the marginal change in image quality per epoch can be inferred. A negative change in the accuracy of $D_{std}$ indicates that, after the $G$ is trained, the discriminator gets 'fooled' by a higher proportion of fake images, indicating the images are more realistic, relative to the prior iteration. As expected, Figure \ref{gan_eval} shows that the GANs under both training schemes fool the discriminator with a higher proportion of images after each training iteration. The larger reduction in the change in the accuracy is greater per iteration under the gradient training scheme however, which demonstrates the value of GI for generator training.

\begin{algorithm}[!ht]
\caption{Gradient trained GAN}
\label{alg:gan_grad}
\begin{algorithmic}[1] 
\STATE Initialise $G$, $D_{std}$, and $D_{grad}$ networks.
\FOR{iteration}
\STATE Generate fake images using $G$ network.
\STATE Generate and store derivatives of gradient discriminator $D_{grad}$ with respect to real images inputs.
\STATE Train conventional discriminator $D_{std}$.
\STATE Overwrite current $D_{grad}$ parameters with parameters $\theta$ of conventional discriminator $D_{std}$.
\STATE Train $G$ to generated images that produce 'real' targets and gradients according to $D_{std}$ and $D_{grad}$ respectively.
\ENDFOR
\STATE \textbf{return} Generated images
\end{algorithmic}
\end{algorithm}

Table \ref{tab:GAN_res} evaluates GANs that have undergone significant training (15,000 iterations) to compare the image quality of generative networks trained with and without GI. The Kullback–Leibler (KL) divergence is averaged for 150,000 images to give an objective indication of the quality of the images generated under each training scheme. The lower KL-divergence of images generated under the gradient trained network indicate it is producing higher quality images. 

\begin{table}[!ht]
\centering
\begin{tabular}{ll}
\toprule
\textbf{Training scheme}  & \textbf{KL-divergence}  \\
\bottomrule
Conventional       & $1022.130  \pm 330.589$       \\
Gradient       & $1004.036  \pm 331.428$      \\
\bottomrule
\end{tabular}
\caption{KL divergence between 150,000 images generated via GANs trained for 15,000 epochs with and without (conventional) GI versus true images.}
\label{tab:GAN_res}
\end{table}


Under this relatively simple example, the difference between images generated under gradient and conventional trained networks does not appear discernible to the humans. One would therefore expect the apparent benefit of training networks with GI to be more readily apparent for comparatively more challenging (i.e. higher dimensional tasks).

\section{GI for early indication in hyper-parameter tuning: A Delta-Min example}
\label{sec:delta_min}

As explained in Section \ref{grad_val}, the value of GI is a function of model error. This insight motivates a novel method for determining an upper bound to NN model complexity, coined delta-min or $\delta_{min}$, where complexity refers to the number of nodes in a feed-forward NN. The central idea is that the diminution in the observable value of GI that results from changes to hyper-parameters can be an early indicator of bounds that hyper-parameters should not cross in optimality.

\subsection{Deriving Delta-Min}
The assumptions behind the delta-min method are as follows: First consider an accuracy measure $A$ to evaluate the performance of a model's predictions. Root mean-squared error (RMSE) is a straightforward choice for regression problems although the following logic generalises for any accuracy measure. The RMSE of a NN's $N$ predictions ($\mathbf{\hat{y}}$) of true target variables ($\mathbf{y}$) is 
\begin{equation}
	\label{rmse}
	A[\hat{y}] = \sqrt{\frac{(\mathbf{y} - \mathbf{\hat{y}})^2)}{N}}.
\end{equation}
A NN's predictions can be expressed as a function of model complexity $\mathbf{\hat{y}} = F_{\varphi}(\mathbf{X}, c)$ where $c$ refers to the number of nodes within a NN and $\mathbf{X}$ refers to a input data. 
  
\textbf{Assumption 1}: \textit{$\frac{\delta A}{\delta c} \in \mathbb{R}$ is a monotonic non-decreasing \footnote{The specific accuracy measure selected alters the specifics of this assumption. For an accuracy measure such as correlation then $\frac{\delta A}{\delta c}$ must be a monotonic non-increasing function as a more accurate model has a \textit{higher} correlation to $\mathbf{y}$ (rather than a lower RMSE).} function of $c$.}
 
There is some optimal model complexity $c^{*}$ that produces the optimal accuracy measure (e.g. lowest RMSE) $A^{*}$. Naturally, a practitioner designing NN architecture seeks to find $c^{*}$. A NN  trained with GI can be denoted as  $F_{\varphi}^{G}(\mathbf{X}, c)$. 
 
\textbf{ Assumption 2}: \textit{For $c \leq c*$: $A \left[ F_{\varphi}^{G}(\mathbf{X}, c) \right] \leq A \left[ F_{\varphi}(\mathbf{X}, c) \right]$ where A is RMSE (invert for accuracy measures that increase as predictions become more accurate, i.e correlation).}

GI has value (improves model accuracy) when model complexity is $\textit{below}$ optimal.
 
\textbf{ Assumption 3}: \textit{For $c \leq c*$: $\nabla_{c} A \left[ F_{\varphi}^{G}(\mathbf{X}, c) \right] \geq \nabla_{c} A \left[ F_{\varphi}(\mathbf{X}, c) \right]$ where A is RMSE (invert for accuracy measures that increase as predictions become more accurate, i.e correlation).}

This strict assumption implies that when models have below optimal complexity, the improvement in accuracy that follows from increasing complexity (Assumption 1) is at least as large for models trained without gradients relative to those trained with. This requires the value of GI for model training to diminish as a model approaches optimal. 


Under these assumptions, the minimum absolute difference between the accuracy of NNs trained with and without GI can be used as an indicator of an upper bound to model complexity $c_{upper}$. This complexity can be found using the delta-min metric:
 \begin{equation}
	\label{delta_min}
	\delta_{min} = \min_{c} |A \left[ F_{\varphi}^{G}(\mathbf{X}, c) \right] - A \left[ F_{\varphi}(\mathbf{X}, c) \right]|.
\end{equation}

To support these assumptions and to demonstrate how they can be utilised to determine $c_{upper}$, $\delta_{min}$ is computed for 3 toy processes with dimensionalities of 2, 4 and 8. 

\subsubsection{Surrogate model for computer simulator}
A 2-dimensional dynamic toy problem ~\cite{chen2019bayesian} is defined to act as a toy computer simulator where only inputs $t$ and outputs $y_t$ can be observed:
\begin{equation}
	\label{eq:2d_func}
	y_t = (\cos [ \phi(\theta_1 - t - \psi)] \cos [ \phi(\theta_2 - t - \psi)])^2,
\end{equation}
where $\phi = 0.1$, $\psi = 5$, and $t$ is the time step ranges between $T = [1, 2,..., 9, 10]$. We construct a NN based surrogate model to learn this simulator for input parameters $\boldsymbol{\theta}_{2D} = [\theta_1, \theta_2]$ drawn from the distribution $\mathbf{\theta} \sim U_{[0, 15]}$. Figure illustration of the toy simulator and the comparison results between the standard and the gradient trained surrogate models can be found in Appendix A.1.

\subsubsection{The GARCH model}
The 4-dimensional problem has the form of a Generalised Autoregressive Conditional Heteroscedasticity (GARCH) model, originally introduced by \cite{BOLLERSLEV1986307}. GARCH models are a popular volatility model for economic and financial time series data  (for seminal examples see \cite{FRENCH19873} and \cite{franses1996forecasting}). The GARCH (1,1) used is:
\begin{equation}
	\label{garch_eq}
	\sigma_{t+1}^2 = \omega + \alpha u_{t}^2 + \beta \sigma_{t}^2.
\end{equation}
This model assumes t+1 volatility, $\sigma_{t+1}^2$, is made of a constant $\omega$, period t forecasted volatility $\sigma^2$ and period t squared return $\mu^2$, weighted with $\alpha$ and $\beta$. In order to add a dynamic component to the toy problem, volatility predictions in future periods $t + h$ (where $h \geq 2$) are computed where subsequent forecasts exploit the relationship $E_t [u_{t+1}^2]= \sigma_{t+1}^2$. We have
\begin{equation}
	\label{garch_th}
	\begin{split}
	\sigma_{t+h}^2 &= \omega + \alpha E_t [u_{t+h-1}^2] + \beta E_t [\sigma_{t+h-1}^2] \ for \ h \geq 2
	\\
	&= \omega + (\alpha + \beta) E_t [\sigma_{t+h-1}^2] \ for \ h \geq 2.
	\end{split}
\end{equation}
The NN model must therefore capture the behaviour of the GARCH(1,1) process described in equations \ref{garch_eq} and \ref{garch_th}. The input parameter set for this example is 4-dimensional, $\boldsymbol{\theta}_{4D} = [\mu, \omega, \alpha, \beta]$. A forecast window of $t + h$ for $h \in [1, 2, 3, 4, 5]$ is considered (e.g. volatility is forecasted up to 5 days in the future). Figures illustrating the performance of the GARCH(1,1) model can be found in Appendix A.2. 


\begin{table}[!h]
 	\centering
 	\small
\begin{tabular}{ |c|c|c|c| } 
\hline
Problem & Training size & Optimal model & $\delta_{min}$ UB \\
\hline
\multirow{5}{1em}{2D} & 50 & 36 & 40 \\ 
& 100 & 40 & 52 \\ 
& 300 & 28 & 52 \\
& 500 & 24 & 32 \\
& 700 & 20 & 24 \\
\hline
\multirow{4}{1em}{4D} & 100 & 10 & 46 \\ 
& 300 & 6 & 22 \\
& 500 & 6 & 18 \\
& 700 & 6 & 14 \\
\hline
\multirow{4}{1em}{8D} & 2500 & 30 & 30 \\ 
& 5000 & 20 & 20 \\
& 7500 & 30 & 30 \\
& 10000 & 20 & 70 \\
\hline
\end{tabular}
 \caption{Optimal model complexities (defined as complexity that produces lowest RMSE on unseen data) compared to the complexity upper-bound implied by the delta-min metric.}
 \label{2d_ub_vs_opt}
 \end{table}

\begin{figure*}[!ht]
	\centering
	\includegraphics[scale=0.5]{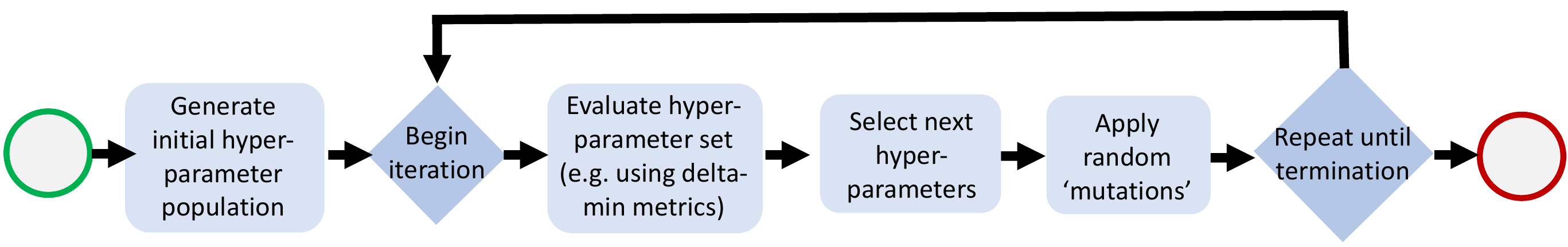}
	\caption{Genetic algorithm optimisation procedure incorporating delta-min metric.} 
	\label{genetic_algo}
\end{figure*}

The 8-dimensional function, shown in Equation \ref{8d_func}, is designed for its relatively high dimensionality and non-linearity.
\begin{equation}
  \label{8d_func}
  \begin{split}
  y = \beta_1 \mathbf{x}_0 + \beta_2(\beta_3  \exp \left[ \beta_4 (\mathbf{x}_1 - \alpha) \right]) + \ 
  \\
  \beta_5(\beta_6  \exp \left[ \beta_7 (\mathbf{x}_2 - \eta) \right]) + \beta_8.
  \end{split}
\end{equation}
Parameters $\alpha$ and $\eta$ are fixed and  $\mathbf{X} = [\mathbf{x_0}, \mathbf{x}_1, \mathbf{x}_2]$, representing known input data with 3 features. The NN model therefore emulates parameters $\boldsymbol{\theta}_{8D} = \{\beta_j\}_{j=1}^8$ over an arbitrarily chosen input region $\mathbf{X}  \sim U_{[0, 1]}$.

The upper-bounds implied using the GI assisted delta-min method for the three example problems are presented in Table \ref{2d_ub_vs_opt}. Results show that the optimal model complexity, defined as the complexity which minimises model RMSE on unseen data for each training sample size consistently lies below the delta-min upper bound. 
 
\subsection{Genetic Algorithms}
\label{genetic_algos}

The proposed value of the delta-min upper bound is that it should allow a practitioner to rule out large model complexities, that are the most computationally expensive to train, early on in a hyper-parameter optimisation pipeline. Genetic algorithms \cite{mitchell1995genetic}, so called because they incorporate biological phenomena such as random mutation and survival of the fittest, are a class of automatic optimisation methods that have become increasingly popular in recent years and natural application for the delta-min metric.

Figure \ref{genetic_algo} demonstrates an example automatic hyper-parameter optimisation pipeline that incorporates the $\delta_{min}$ metric to evaluate model complexity. 

Model complexities above the complexity implied by $\delta_{min}$ can be ruled out in early iterations of the algorithm allowing for more efficient evaluation of granular hyper-parameter values in subsequent runs. Conducting experiments validating the utility of the $\delta_{min}$ metric for hyper-parameter optimisation using genetic algorithms is a recommended area for future research. 

\section{Ridge-Gradients Regularisation}
\label{rg_reg}
To demonstrate the value of gradients for improving linear regression model fit this work presents a novel regularisation method coined Ridge-Gradients (RG) regularisation. The method builds upon Ridge regularisation, a well known modification to ordinary least squares (OLS) linear regression originally proposed bv  \cite{hoerl1970ridge} that penalises large coefficient values. Regularised linear models often outperform non-regularised equivalents as large coefficient values are typically a feature of models which have overfit to training data and therefore generalise poorly to unseen data.

The Ridge-Gradients optimisation problem is defined as:
\begin{equation}
	\label{rg_opt}
\min_{\hat{\boldsymbol{\beta_{rg}}}} \ \ (\mathbf{z} -\mathbf{\hat{z}})^2 + \lambda_1 \hat{\boldsymbol{\beta}}_{rg}^2 + \lambda_2 (\boldsymbol{\hat{\beta}_{rg}} - \boldsymbol{\hat{\beta}_{grad}})^2.
\end{equation}
This optimisation problem includes the conventional goal of minimising the difference between model predictions $\mathbf{\hat{z}}$ and true process outputs $\mathbf{z}$ and penalises large coefficient values with the standard Ridge shrinkage parameter $\lambda_1 \geq 0$. Gradients are leveraged in the third term which involves penalising, using $\lambda_2 \geq 0$, the difference between the coefficient   $\boldsymbol{\hat{\beta}_{rg}}$ and a new quantity, $ \boldsymbol{\hat{\beta}_{grad}}$. $\boldsymbol{\hat{\beta}_{grad}}$ is the coefficient vector calculated using OLS with Ridge regularisation that fits \textit{derivative inputs} to output gradients $\delta \mathbf{z}/\delta \mathbf{x}$.

\begin{figure}[hbt!]
\centering
\hbox{\hspace{-2em}\includegraphics[scale=0.5]{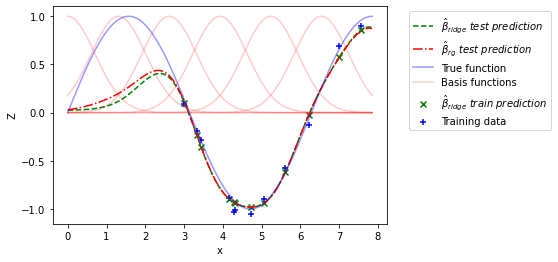}}
\caption{OLS model predictions of sine wave using 12 training data samples and 7 evenly distributed basis functions with r=1 (and a constant).  $\boldsymbol{\hat{\beta}_{ridge}}$ is calculated using OLS with Ridge regularisation with $\lambda_1 = 0.1$ optimised using a validation data set. $\boldsymbol{\hat{\beta}_{rg}}$ is calculated using OLS with Ridge-Gradients regularisation with $\lambda_1 = 0.1$ and $\lambda_2 = 0.1$ optimised using a validation data set. Test RMSE of $\boldsymbol{\hat{\beta}_{rg}}$ and $\boldsymbol{\hat{\beta}_{ridge}}$ is 0.31 and 0.35 respectively (compared to a standard deviation of the test output data of 0.69)} 
\label{sine_wave}
\end{figure}

To understand $\boldsymbol{\hat{\beta}_{grad}}$, consider the following example where the true process is $P(\mathbf{x}) = \sin{\mathbf{x}}$ and $\mathbf{x}$ is 1-dimensional input vector over the interval $[0, 2 \pi]$. To fit a linear model to a non-linear sine wave the 1-dimensional input vector $\mathbf{x}$ can be projected in to higher dimensional space using radial basis functions (RBF) with evenly spaced centres in the input interval. This is illustrated in Figure  \ref{sine_wave} and the RBF equation is shown in  Equation \ref{RBF}. The resulting input matrix is denoted $\mathbf{\Phi(x)}$ and has dimensionality $N$ x $(C + 1)$ where $N$ is the size of the training sample and $C$ is the number of basis functions used plus a constant term. This example proceeds with $C = 7$ basis functions as this is sufficient to fit a sine curve reasonably well \footnote{The exact number of basis functions does not effect the key results and was determined via trial and error. See \cite{tipping2003fast} for automatic methods to optimally determine this quantity.}.

\begin{equation}
	\label{RBF}
	\Phi_i(\mathbf{x}) = \exp \left \{ \frac{-(\mathbf{x} - c_i)^2}{r^2} \right \} 
\end{equation}

The linear model with coefficients $ \boldsymbol{\hat{\beta}_{ridge}} = [\hat{\beta_0}_{ridge},  \hat{\beta_1}_{ridge}, ... , \hat{\beta_7}_{ridge}]^T$ fit using OLS with ridge regularisation is:

\begin{equation}
	\begin{split}
	\label{lin_model}
	\hat{\mathbf{z}} &= \boldsymbol{\hat{\beta}_{ridge}} \ \boldsymbol{\Phi(x)}  = \hat{\beta_0}_{ridge} + \ 
	\\
	& \hat{\beta_1}_{ridge} \ \exp \left \{ \frac{-(\mathbf{x} - c_1)^2}{r^2} \right \} + ... \ 
	\\
	&... + \hat{\beta_7}_{ridge} \exp \left \{  \frac{-(\mathbf{x} - c_7)^2}{r^2}\right \}
	\end{split}
\end{equation}

In this example both necessary conditions required to leverage GI are satisfied as the derivative of $\mathbf{z}$ with respect to $\mathbf{x}$ is easily calculated before model learning and a linear regression output $\mathbf{\hat{z}}$ is differentiable with respect to its inputs $\mathbf{x}$. After calculating and retaining $\frac{\delta \mathbf{z}}{\delta \mathbf{x}}$, OLS (again with ridge regularisation) can be used to calculate $\boldsymbol{\hat{\beta}_{grad}} =  [\hat{\beta_0}_{grad},  \hat{\beta_1}_{grad}, ... , \hat{\beta_7}_{grad}]^T$, training the derivative inputs to the derivative outputs. The resulting 'gradient' fitting model takes the form:

\begin{equation}
	\label{beta_grad}
	\begin{split}
\frac{\delta \mathbf{\hat{z}}}{\delta \mathbf{x} } &= \hat{\beta_0}_{grad} +
 \hat{\beta_1}_{grad} \frac{-2(\mathbf{x}  - c_1)^2}{r^2} \exp \left \{ \frac{-(\mathbf{x} - c_1)^2}{r^2} \right \} + \
\\
&... + \hat{\beta_7}_{grad} \frac{-2(\mathbf{x} - c_7)^2}{r^2} \exp \left \{ \frac{-(\mathbf{x}  - c_7)^2}{r^2} \right \}
	\end{split}
\end{equation}

The derivative of original OLS model from Equation \ref{lin_model} with respect to inputs can also be calculated to obtain:

\begin{equation}
	\label{beta_ridge}
	\begin{split}
		\frac{\delta \mathbf{\hat{z}}}{\delta \mathbf{x}} &= \hat{\beta_0}_{ridge} + \hat{\beta_1}_{ridge} \frac{-2(\mathbf{x} - c_1)^2}{r^2} \exp \left \{ \frac{-(\mathbf{x} - c_1)^2}{r^2} \right \} + \\
		&... + \hat{\beta_7}_{ridge} \frac{-2(\mathbf{x} - c_7)^2}{r^2} \exp \left \{ \frac{-(\mathbf{x} - c_7)^2}{r^2} \right \}
	\end{split}
\end{equation}

The key difference between $\boldsymbol{\hat{\beta}_{grad}}$ and $\boldsymbol{\hat{\beta}_{ridge}}$ is that $\boldsymbol{\hat{\beta}_{grad}}$ is \textit{fit} using OLS with the gradients of outputs with respect to inputs $\frac{\delta \mathbf{z}}{\delta \mathbf{x}}$ as the target variable, while $\boldsymbol{\hat{\beta}_{ridge}}$ uses outputs $\mathbf{z}$. Naturally this means $\boldsymbol{\hat{\beta}_{grad}}$ will be a better predictor of the true \textit{gradients} of the underlying function than $\boldsymbol{\hat{\beta}_{ridge}}$. This is illustrated in the Figure presented in the supplementary materials, Technical Appendix B.2.

From Equations \ref{beta_grad} and \ref{beta_ridge} it becomes clear that in an optimal model there would be no difference between  $\boldsymbol{\hat{\beta}_{grad}}$ and $\boldsymbol{\hat{\beta}_{ridge}}$ in order to optimally predict $\frac{\delta \mathbf{z}}{\delta \mathbf{x}}$. The goal of the additional regularisation term shown in Equation \ref{rg_opt} is therefore to penalise differences between the coefficient vector fit to the target variable and the coefficient vector that optimally fits the target's gradients as in an optimal model, both should be fit well.
\begin{equation}
\label{beta_rg}
\boldsymbol{\hat{\beta}_{rg}} = (\mathbf{\Phi}'\mathbf{\Phi} + \lambda_1 \mathbf{I} + \lambda_2 \mathbf{I})^{-1} (\lambda_2 \ \boldsymbol{\hat{\beta}_{grad}} + \mathbf{\Phi}'\mathbf{Z})
\end{equation}

\begin{figure}[!ht]
	\centering
	\includegraphics[scale=0.55]{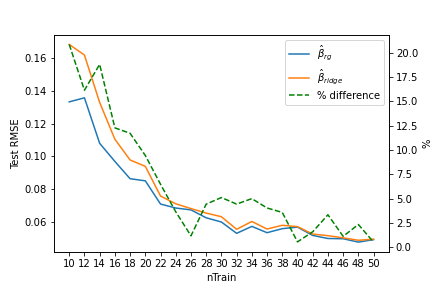}
	\caption{Comparative performance of $\boldsymbol{\hat{\beta}_{ridge}}$ and $\boldsymbol{\hat{\beta}_{rg}}$ over 50 random samples of training data of sizes in intervals of 2 from 10 to 50. Positive percentage difference indicates $\boldsymbol{\hat{\beta}_{rg}}$ outperforms.} 
	\label{beta_versus_betarg}
\vspace{-0.4cm}
\end{figure}

The RG solution is shown in Equation \ref{beta_rg} (full derivation shown in supplementary materials, Technical Appendix B.1):


This regularisation method consistently outperforms vanilla Ridge regularisation over a range of training samples and sizes, as shown in Figure \ref{beta_versus_betarg}.

\section{Conclusion}
\label{Con}
This paper presents methods for leveraging GI to improve model prediction, focussing on applications for GANs, hyper-parameter tuning and linear regression regularisation. The results demonstrate that the intrinsic GI extracted from the original model and the training data can be used to improve the performance of machine learning models. This approach is suitable for problems where the computer model (e.g., numerical simulators, generator networks in GAN) or mathematical expressions of the underlying (physical) process is known and differentiable. The proposed method is particularly suitable for problems with very limited available training samples since the extracted GI can be viewed as extra training samples under certain conditions.  

\bibliographystyle{named}
\bibliography{ijcai22}

\begin{thebibliography}{}

\bibitem[\protect\citeauthoryear{Bollerslev}{1986}]{BOLLERSLEV1986307}
Tim Bollerslev.
\newblock Generalized autoregressive conditional heteroskedasticity.
\newblock {\em Journal of Econometrics}, 31(3):307--327, 1986.

\bibitem[\protect\citeauthoryear{Brehmer \bgroup \em et al.\egroup
  }{2020}]{Brehmer5242}
Johann Brehmer, Gilles Louppe, Juan Pavez, and Kyle Cranmer.
\newblock Mining gold from implicit models to improve likelihood-free
  inference.
\newblock {\em Proceedings of the National Academy of Sciences},
  117(10):5242--5249, 2020.

\bibitem[\protect\citeauthoryear{Chen and Hobson}{2019}]{chen2019bayesian}
Xi~Chen and Mike Hobson.
\newblock Bayesian surrogate learning in dynamic simulator-based regression
  problems, 2019.

\bibitem[\protect\citeauthoryear{Deng}{2012}]{deng2012mnist}
Li~Deng.
\newblock The mnist database of handwritten digit images for machine learning
  research.
\newblock {\em IEEE Signal Processing Magazine}, 29(6):141--142, 2012.

\bibitem[\protect\citeauthoryear{Franses and
  Van~Dijk}{1996}]{franses1996forecasting}
Philip~Hans Franses and Dick Van~Dijk.
\newblock Forecasting stock market volatility using (non-linear) garch models.
\newblock {\em Journal of Forecasting}, 15(3):229--235, 1996.

\bibitem[\protect\citeauthoryear{French \bgroup \em et al.\egroup
  }{1987}]{FRENCH19873}
Kenneth~R. French, G.William Schwert, and Robert~F. Stambaugh.
\newblock Expected stock returns and volatility.
\newblock {\em Journal of Financial Economics}, 19(1):3--29, 1987.

\bibitem[\protect\citeauthoryear{Goodfellow \bgroup \em et al.\egroup
  }{2014}]{goodfellow2014generative}
Ian~J. Goodfellow, Jean Pouget-Abadie, Mehdi Mirza, Bing Xu, David
  Warde-Farley, Sherjil Ozair, Aaron Courville, and Yoshua Bengio.
\newblock Generative adversarial networks, 2014.

\bibitem[\protect\citeauthoryear{Gutmann \bgroup \em et al.\egroup
  }{2018}]{gutmann2018likelihood}
Michael~U Gutmann, Ritabrata Dutta, Samuel Kaski, and Jukka Corander.
\newblock Likelihood-free inference via classification.
\newblock {\em Statistics and Computing}, 28(2):411--425, 2018.

\bibitem[\protect\citeauthoryear{Hoerl and Kennard}{1970}]{hoerl1970ridge}
Arthur~E Hoerl and Robert~W Kennard.
\newblock Ridge regression: Biased estimation for nonorthogonal problems.
\newblock {\em Technometrics}, 12(1):55--67, 1970.

\bibitem[\protect\citeauthoryear{Mitchell}{1995}]{mitchell1995genetic}
Melanie Mitchell.
\newblock Genetic algorithms: An overview.
\newblock {\em Complexity}, 1(1):31--39, 1995.

\bibitem[\protect\citeauthoryear{Tipping and Faul}{2003}]{tipping2003fast}
Michael~E Tipping and Anita~C Faul.
\newblock Fast marginal likelihood maximisation for sparse bayesian models.
\newblock In {\em International workshop on artificial intelligence and
  statistics}, pages 276--283. PMLR, 2003.

\bibitem[\protect\citeauthoryear{Tran \bgroup \em et al.\egroup
  }{2017}]{tran2017hierarchical}
Dustin Tran, Rajesh Ranganath, and David~M Blei.
\newblock Hierarchical implicit models and likelihood-free variational
  inference.
\newblock In {\em Proceedings of the 31st International Conference on Neural
  Information Processing Systems}, pages 5529--5539, 2017.

\end{thebibliography}

\clearpage
\appendix

\section{Section 4 supplementary material}

\subsection{Comparison between conventional and gradient trained models in 2D toy problem}

\begin{figure}[!ht]
	\centering
	\includegraphics[scale=0.28]{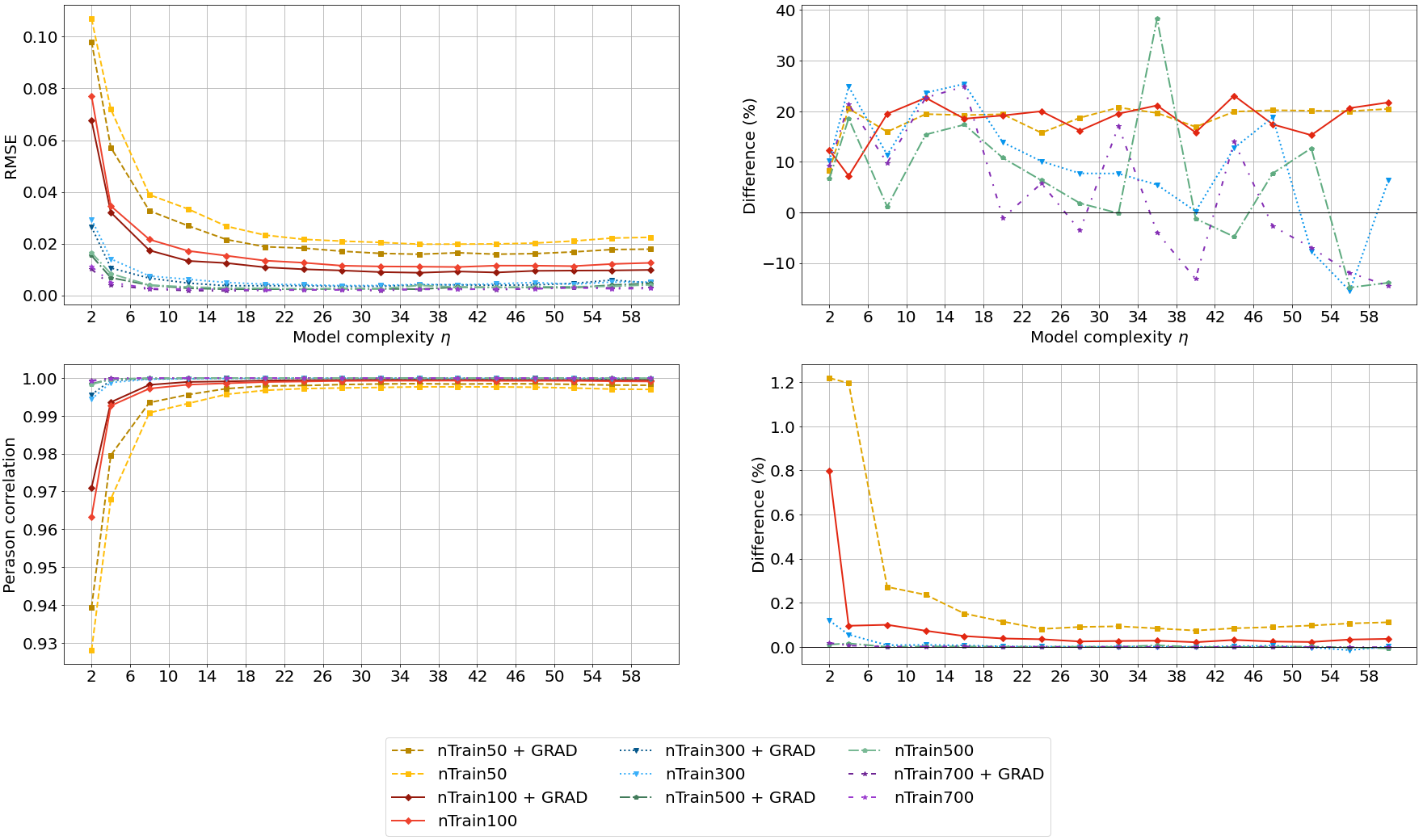}
	\caption{2D problem surrogate model test accuracy. RMSE left)/Pearson correlation (right) of surrogates models prediction on 1000 test data samples. Surrogates are trained on varying training sample sizes with varying complexities and either with (darker colours) or without (lighter colours) gradients. Model complexity values $\eta$ correspond to a NN with 4 hidden layers, each with $\eta * 4$ nodes. Results are averaged over 10 random draws from a fixed pool of 1000 synthetic training data points generated by the cosine function. Axes are set to emphasises region of interest.}
	\label{2d_complexity}
\end{figure}

\clearpage
\subsection{Performance of GARCH(1,1) model on S\&P data}

\begin{figure}[!h]
	\centering
	\includegraphics[scale=0.55]{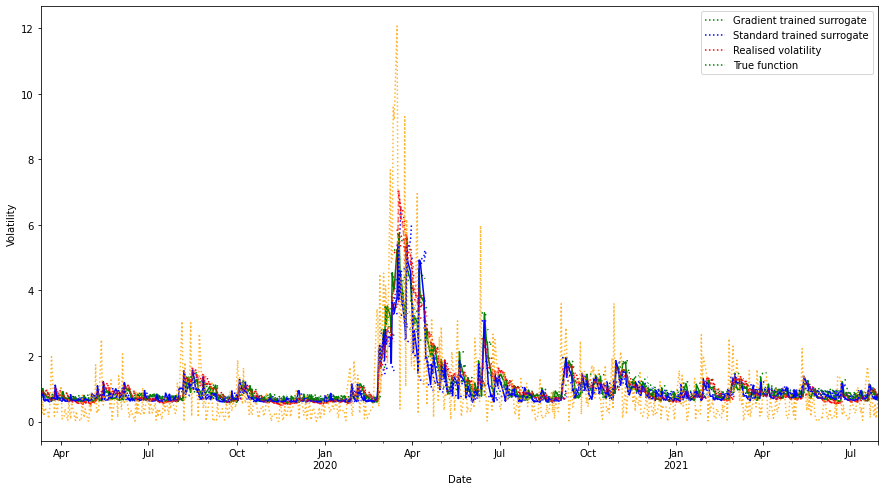}
	\caption{Rolling volatility ($\sigma^2$) prediction of an S\&P 500 test series using optimal models trained with and without gradients on 100 data points and the true GARCH function. T+1 predictions are connected with a coloured line and t+2 to t+5 predictions are shown projecting from this line. Latent parameters are re-estimated each day using maximum likelihood estimation using the prior 200 days of return data. Average of 5 surrogate models (each trained on random samples of training data for each sample size) is used for predicted values.}
	\label{GARCH_pred}
\end{figure}

\clearpage
\section{Section 5 supplementary material}
\subsection{Ridge-Gradients Derivation}
\label{rg_derivation}
This appendix provides the derivation of the Ridge-Gradients coefficient shown in Equation 10 from Section 3.1. Transposed matrices and vectors denoted with a superscript $T$, e.g $\mathbf{z}^T$ and the identity matrix denoted $\mathbf{I}$.

The objective function is:

\begin{equation}
	\label{rg_opt2}
	\min_{\hat{\boldsymbol{\beta_{rg}}}} \ \ Obj = (\mathbf{z} -\mathbf{\hat{z}})^2 + \lambda_1 \hat{\boldsymbol{\beta}}_{rg}^2 + \lambda_2 (\boldsymbol{\hat{\beta}_{rg}} - \boldsymbol{\hat{\beta}_{grad}})^2
\end{equation}

Let $\mathbf{\hat{z}} = \mathbf{X}  \hat{\boldsymbol{\beta}}_{rg}$ where $\mathbf{X}$ is a matrix containing the independent variable data with dimensionality $N$ x $M$ and $\hat{\boldsymbol{\beta}}_{rg}$ is the Ridge-Gradients coefficient vector with dimensionality $M$ x $1$.  The objective function can therefore be written as:

\begin{equation}
	\label{rg_opt3}
	Obj = (\mathbf{z} -\mathbf{X}  \hat{\boldsymbol{\beta}}_{rg})^2 + \lambda_1 \hat{\boldsymbol{\beta}}_{rg}^2 + \lambda_2 (\boldsymbol{\hat{\beta}_{rg}} - \boldsymbol{\hat{\beta}_{grad}})^2
\end{equation}

Expanding this:

\begin{equation}
	\label{rg_opt_expand}
	\begin{split}
	Obj = \mathbf{z}^T \mathbf{z} -  \hat{\boldsymbol{\beta}}_{rg}^T \mathbf{X}^T \mathbf{z} - \mathbf{z}^T \mathbf{X}  \hat{\boldsymbol{\beta}}_{rg} + \hat{\boldsymbol{\beta}}_{rg}^T \mathbf{X}^T \mathbf{X} \hat{\boldsymbol{\beta}}_{rg}  
	\\
	+ \lambda_1 \hat{\boldsymbol{\beta}}_{rg}^T \hat{\boldsymbol{\beta}}_{rg} + \lambda_2 \hat{\boldsymbol{\beta}}_{rg}^T \hat{\boldsymbol{\beta}}_{rg} - \lambda_2  \hat{\boldsymbol{\beta}}_{rg}^T \hat{\boldsymbol{\beta}}_{grad} 
	\\
	- \lambda_2  \hat{\boldsymbol{\beta}}_{grad}^T \hat{\boldsymbol{\beta}}_{rg} + \lambda_2  \hat{\boldsymbol{\beta}}_{grad}^T \hat{\boldsymbol{\beta}}_{grad}
\end{split}
\end{equation}

This can be simplified to:

\begin{equation}
	\label{rg_opt_simp}
	\begin{split}
		Obj = \mathbf{z}^T \mathbf{z}  - 2 \mathbf{z}^T \mathbf{X}  \hat{\boldsymbol{\beta}}_{rg} + \hat{\boldsymbol{\beta}}_{rg}^T \mathbf{X}^T \mathbf{X} \hat{\boldsymbol{\beta}}_{rg} + 
		\\
		 \lambda_1 \hat{\boldsymbol{\beta}}_{rg}^T \hat{\boldsymbol{\beta}}_{rg} + \lambda_2 \hat{\boldsymbol{\beta}}_{rg}^T \hat{\boldsymbol{\beta}}_{rg} - 2 \lambda_2  \hat{\boldsymbol{\beta}}_{rg}^T \hat{\boldsymbol{\beta}}_{grad} + 
		 \\
		 \lambda_2  \hat{\boldsymbol{\beta}}_{grad}^T \hat{\boldsymbol{\beta}}_{grad}
	\end{split}
\end{equation}

Taking to the derivative and imposing the necessary condition to minimise the objective function:

\begin{equation}
	\label{obj_derivative}
	\begin{split}
		\frac{\delta Obj}{\delta \hat{\boldsymbol{\beta}}_{rg}} =  -2\mathbf{X}^T \mathbf{z} + 2 \mathbf{X}^T \mathbf{X} \hat{\boldsymbol{\beta}}_{rg} + 
		\\
		2  \lambda_1 \hat{\boldsymbol{\beta}}_{rg} +
		2 \lambda_2 \hat{\boldsymbol{\beta}}_{rg} - 2 \lambda_2  \hat{\boldsymbol{\beta}}_{grad} = 0
	\end{split}
\end{equation}

Which can be simplified to: 

\begin{equation}
	\label{obj_derivative_simp}
	\begin{split}
	\frac{\delta Obj}{\delta \hat{\boldsymbol{\beta}}_{rg}} =  -\mathbf{X}^T \mathbf{z} + \mathbf{X}^T \mathbf{X} \hat{\boldsymbol{\beta}}_{rg} +  \lambda_1 \hat{\boldsymbol{\beta}}_{rg} +
	\\
	\lambda_2 \hat{\boldsymbol{\beta}}_{rg} - \lambda_2  \hat{\boldsymbol{\beta}}_{grad} = 0
	\end{split}
\end{equation}

Rearranging terms:

\begin{equation}
	\label{obj_derivative_rearange}
	\hat{\boldsymbol{\beta}}_{rg}(\mathbf{X}^T\mathbf{X} + \lambda_1 \mathbf{I} + \lambda_2 \mathbf{I}) = \lambda_2  \hat{\boldsymbol{\beta}}_{grad} +  \mathbf{X}^T \mathbf{z}
\end{equation}

Therefore the solution for $\hat{\boldsymbol{\beta}}_{rg}$ can be written as:

\begin{equation}
	\label{beta_rg_final}
	\hat{\boldsymbol{\beta}}_{rg} = (\mathbf{X}^T\mathbf{X} + \lambda_1 \mathbf{I} + \lambda_2 \mathbf{I})^{-1}(\lambda_2  \hat{\boldsymbol{\beta}}_{grad} +  \mathbf{X}^T \mathbf{z})
\end{equation}

\clearpage
\subsection{\texorpdfstring{$\boldsymbol{\hat{\beta}_{grad}}$}{Lg} predictions of target gradients}

\begin{figure}[!h]
	\centering
	\includegraphics[width=12cm]{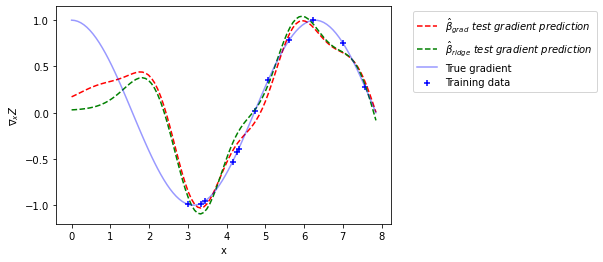}
	\caption{Predictions of the gradient of the sine wave using $\boldsymbol{\hat{\beta}_{grad}}$ and $\boldsymbol{\hat{\beta}_{ridge}}$. The RMSE using $\boldsymbol{\hat{\beta}_{ridge}}$ is 0.4 and the RMSE using $\boldsymbol{\hat{\beta}_{grad}}$ is 0.37.} 
	\label{gradient_pred}
\end{figure}

\end{document}